%% file: main.tex
\documentclass[conference]{IEEEtran}
\usepackage{cite}
\usepackage{amsmath,amssymb,amsfonts}
\usepackage{booktabs}
\usepackage{multirow}
\usepackage{graphicx}
\usepackage[caption=false,font=footnotesize]{subfig}
\usepackage{xcolor}
\usepackage{url}
\usepackage{listings}
\usepackage{microtype}

\begin{document}

\title{L2GSSL: Label-Efficient Local-to-Global Self-Supervised Learning for Volumetric Healthcare Data}

\author{
\IEEEauthorblockN{Po-Cheng Cheng}
\IEEEauthorblockA{
D3 Center, The University of Osaka, Japan\\
po-cheng@is.ids.osaka-u.ac.jp
}
\and
\IEEEauthorblockN{Hung-Hsuan Chen}
\IEEEauthorblockA{
Department of Computer Science and Information Engineering,\\
National Central University, Taiwan\\
hhchen1105@acm.org
}
}

\maketitle
\input{content/abs}

\begin{IEEEkeywords}
self-supervised learning, healthcare data, 3D medical imaging, computed tomography, representation learning
\end{IEEEkeywords}

\input{content/intro}
\input{content/rel-work}
\input{content/method}
\input{content/exp}
\input{content/disc}
\input{content/conc}

\section*{Acknowledgment}
This work was supported by the National Science and Technology Council of Taiwan under Grant 113-2221-E-008-100-MY3. Computational and storage resources were provided by the National Center for High-Performance Computing of the National Applied Research Laboratories, Taiwan.
The authors used LLMs (e.g., Claude and ChatGPT) to improve language and readability. The authors reviewed and edited the content as needed. The authors take full responsibility for the content.

\bibliographystyle{IEEEtran}
\bibliography{ref}
\end{document}

%% file: content/abs.tex
\begin{abstract}
Healthcare institutions routinely accumulate high-dimensional computed tomography (CT) volumes, but only a fraction can be annotated by experts. We present L2GSSL, a local-to-global self-supervised learning framework that leverages unlabeled volumes by decomposing pretraining into intra-slice and inter-slice representation spaces. A slice encoder learns view-invariant local features, and a global encoder aggregates the resulting tokens to model volumetric context. The stop-gradient isolates the local space from the global objective, while slice-wise execution and a three-phase schedule support training under constrained GPU memory. On LIDC-IDRI lung nodule malignancy classification, L2GSSL reaches $0.9229$ AUROC with only $10\%$ of the labels and $0.9377$ with $25\%$, exceeding the volume-level VICReg by $0.0229$ and $0.0294$, respectively. With all labels, the strongest supervised baseline exceeds L2GSSL by only $0.0044$ AUROC, consistent with self-supervised pretraining providing its greatest value when annotations are limited. A model fine-tuned on $25\%$ of LIDC-IDRI and evaluated on the external LNDb cohort without further training obtains $0.8138$ AUROC and $0.7589$ AUPRC, improving over the strongest evaluated self-supervised baseline by $0.1130$ and $0.0954$. Experiments across data sources, thoracic and abdominal anatomy, downstream tasks, and two backbones examine healthcare-data variety, while masking malignant lesion-bearing slices at inference sharply reduces performance, indicating that predictions rely substantially on lesion content. We release the code for reproducibility.\footnote{\url{https://github.com/ykmcheng/L2GSSL}}
\end{abstract}

%% file: content/intro.tex
\section{Introduction}
Healthcare institutions routinely accumulate archives of computed tomography (CT) scans, but expert-curated labels remain available for only a fraction of those data. Each scan is also a high-dimensional volumetric object that combines fine pathological appearance, organ-scale structure, and relationships across ordered slices. The resulting challenge is how to extract useful representations from abundant unlabeled volumes when downstream annotations and accelerator memory are limited. Healthcare archives also vary across institutions, anatomy, acquisition, and annotation practice, so representation quality must be assessed beyond a single internal dataset.

Self-supervised learning (SSL) addresses annotation scarcity by learning from the images themselves. Existing medical SSL approaches include restoration-based pretext tasks~\cite{chen2019context,zhou2021genesis}, contrastive objectives, and volumetric spatial objectives such as VoCo~\cite{wu2024voco}. However, applying a single objective to the final volume representation does not explicitly acknowledge the hierarchy of CT data. An individual slice contains texture, boundary, and local lesion evidence, while the ordered slice sequence carries anatomical continuity and volumetric context. Treating both as a single optimization space may allow a global loss to overwrite useful slice-level structure. Fully 3D pipelines also make it difficult to exploit widely available two-dimensional pretrained encoders and can exceed the resources available to smaller research or clinical sites.

We therefore propose Local-to-Global Self-Supervised Learning (L2GSSL), a structured pretraining framework that learns from unlabeled 3D medical images. It independently regularizes local slice representations and global volume representations using VICReg~\cite{bardes2022vicreg}. A stop-gradient boundary makes the dependency directional: local embeddings serve as inputs to the global encoder during the forward pass, while the global objective does not update the local encoder during backpropagation. A three-phase optimization procedure first establishes slice-level features, then learns both spaces, and finally consolidates inter-slice modeling. Slice-wise processing and feature accumulation avoid keeping an entire end-to-end 3D gradient graph in memory, making hierarchical pretraining feasible within a constrained GPU-memory budget.

We evaluate whether L2GSSL turns unlabeled volumetric data into useful initialization when expert labels are scarce. Pulmonary-nodule malignancy experiments on LIDC-IDRI~\cite{armato2011lidc} use $10\%$, $25\%$, and $100\%$ of the downstream labels, while direct evaluation on LNDb~\cite{pedrosa2019lndb} tests transfer across data sources without target-domain adaptation. A second backbone, lesion-slice masking, and integration into an abdominal-trauma system using RSNA~\cite{hermans2025rsna} and AbdomenCT-1K~\cite{ma2021abdomenct} expand the evaluation across architectures and tasks. Our contributions are as follows.

\begin{itemize}
\item We propose a hierarchical self-supervised learning framework that learns intra-slice and inter-slice representations through separate objectives and directional gradient flow.
\item We develop a resource-conscious training procedure that combines slice-wise execution with phased local and global optimization.
\item We demonstrate that the largest gains occur with $10\%$ and $25\%$ of the downstream labels and diminish with full supervision, matching the intended role of self-supervised pretraining under limited expert annotation.
\item We evaluate transfer and model behavior through direct LIDC-IDRI-to-LNDb evaluation, thoracic and abdominal CT tasks, two backbone designs, radiologist variability, and lesion dependence.
\end{itemize}

%% file: content/rel-work.tex
\section{Related Work}
\subsection{Volumetric Medical Image Learning}
Three-dimensional CNNs such as 3D U-Net~\cite{cicek20163d} and V-Net~\cite{milletari2016vnet} established direct volumetric processing, while transformer-based systems such as UNETR~\cite{hatamizadeh2022unetr} model longer-range dependencies. These architectures are effective, but operate on high-dimensional inputs. An alternative is to encode slices and then aggregate their embeddings. The Medical Slice Transformer (MST) combines a DINOv2-initialized image encoder~\cite{oquab2024dinov2} with inter-slice modeling~\cite{muller2025mst}; this separation aligns naturally with the structure of CT and allows reuse of strong image-level priors. MaxViT~\cite{tu2022maxvit} offers a second hybrid convolution-attention hierarchy through which the portability of a local-to-global objective can be tested.

Recent work adapts volumetric contrastive learning to individual abdominal CT slices and subsequently aggregates pretraining features with a sequential model~\cite{chiu2026volumes}. This 2D-VoCo formulation reduces the burden of direct 3D pretraining while learning slice-level spatial-semantic features. L2GSSL follows the same resource-conscious direction but addresses a different representation problem: it assigns independent objectives to both the slice space and the ordered inter-slice space, rather than restricting self-supervision to the local encoder before downstream sequence modeling.

These two architectural families make different trade-offs. Direct 3D models expose the network to local appearance and volumetric context simultaneously, but their activation memory grows with all three spatial dimensions. Slice-sequence models instead compress each image plane before learning across an ordered token sequence. This 2.5D dataflow preserves through-plane context without applying every operation to a dense 3D tensor and provides a natural interface to image encoders pretrained on large natural-image collections.

L2GSSL differs from direct 3D architectures because its pretraining objectives do not operate only on a dense volumetric representation. Slices are encoded in groups and reconstructed as an ordered embedding sequence for global modeling, preserving compatibility with image-level pretrained encoders. L2GSSL also differs from conventional encode-then-aggregate pipelines: although these pipelines separate forward computation into slice encoding and sequence aggregation, they do not necessarily separate the corresponding optimization objectives. A loss attached only to the final volume embedding can still propagate through and reshape the slice encoder. L2GSSL explicitly regularizes both representation spaces and prevents the global objective from updating the local encoder. It is therefore a pretraining strategy for hierarchical volumetric models rather than an alternative downstream backbone. In this study, we instantiate L2GSSL with MST and MaxViT to evaluate the framework across an explicit slice-sequence architecture and a staged convolution-attention backbone.

\subsection{Self-Supervised Learning for Medical Images}
Early medical SSL approaches reconstruct transformed inputs or predict spatial context. Models such as Genesis apply restoration transformations to 3D images~\cite{zhou2021genesis}, and context restoration learns visual representations without labels~\cite{chen2019context}. Contrastive approaches align augmented views, but often depend on negative samples. In ordered medical volumes, anatomically similar neighboring slices are problematic negatives. SimSiam avoids negative pairs through a predictor and stop-gradient~\cite{chen2021simsiam}; MoCo v2 uses a momentum encoder and queue~\cite{chen2020moco}; VICReg prevents collapse by regularizing invariance, variance, and covariance~\cite{bardes2022vicreg}. VoCo instead learns spatial consistency from overlapping 3D crops~\cite{wu2024voco}. L2GSSL differs in where the objective is applied: it creates separate local and global SSL spaces and controls the information flow between them.

Contrastive and joint-embedding methods learn representations by enforcing agreement between augmented views. Their behavior depends on how positive pairs are constructed, whether and how negative samples are used, and the spatial unit being embedded. This distinction is important for CT because adjacent slices are separate inputs yet may contain nearly identical anatomical structures. Non-contrastive objectives avoid explicitly pushing such samples apart. VICReg is particularly suitable for hierarchical training because it prevents representation collapse through batch statistics without a momentum encoder, memory queue, or predictor. We consequently use the same objective family at both spatial levels, while keeping their projectors, inputs, and gradient paths distinct.

Table~\ref{tab:ssl-design} compares the design characteristics of the SSL baselines evaluated and the related volumetric approaches. MoCo v2, SimSiam, volume-level VICReg, and L2GSSL are evaluated within the controlled MST setting. We include VoCo experimentally as a resource-constrained reference with a different 3D backbone. Among them, L2GSSL applies independent losses to slice and volume embeddings while preventing the volume loss from updating the slice encoder.

\begin{table*}[t]
\centering
\caption{Design comparison of the evaluated SSL baselines and closely related volumetric pretraining approaches. ``Local/global split'' denotes separate objectives and controlled gradient paths, not merely a hierarchical backbone. ``Auxiliary components'' denotes training-only modules beyond the encoder and projector, such as momentum encoders, memory queues, predictors, or EMA teachers.}
\label{tab:ssl-design}
\resizebox{\textwidth}{!}{\begin{tabular}{lccccc}
\toprule
Method & Pretraining input & Alignment target & Explicit negatives & Auxiliary components & Local/global split\\
\midrule
MoCo v2~\cite{chen2020moco} & Slice-aggregated volume & Final volume embedding & Yes & Momentum encoder and queue & No\\
SimSiam~\cite{chen2021simsiam} & Slice-aggregated volume & Final volume embedding & No & Predictor & No\\
Volume VICReg~\cite{bardes2022vicreg} & Slice-aggregated volume & Final volume embedding & No & None & No\\
VoCo~\cite{wu2024largescale} & Overlapping 3D crops & Volumetric spatial relation & Spatially defined & EMA teacher & No\\
2D-VoCo~\cite{chiu2026volumes} & Individual CT slices & Slice-level spatial-semantic relation & Spatially defined & None & No\\
L2GSSL & Slices and ordered tokens & Slice and volume embeddings & No & None & Yes\\
\bottomrule
\end{tabular}}
\end{table*}

Beyond self-supervised representation learning, L2GSSL is related to decoupled learning methods that partition a network into modules optimized by local objectives. Associated Learning, DeInfoReg, and SCPL reduce dependence on a single end-to-end gradient path through local objectives~\cite{wu2022associated, huang2025deinforeg, ho26scpl}. L2GSSL applies the decoupling principle to the spatial hierarchy of volumetric medical images: local and global modules use separate self-supervised objectives, and the global objective is prevented from updating the local encoder.

\subsection{External Evaluation and Model-Behavior Analysis}
Medical-imaging reporting guidelines emphasize the evaluation of independent datasets to assess sensitivity to changes in acquisition source and annotation practice~\cite{tejani2024claim}. External evaluation therefore provides evidence distinct from another internal split, although it does not establish prospective clinical validity.

Model-behavior analyses require similar care. Perturbation tests examine whether predictions change when selected image content is removed, providing evidence of input dependence without establishing causal reasoning or clinical utility.

%% file: content/method.tex
\section{Method}
\subsection{Hierarchical Representation Spaces}
Let a CT volume be $V=[s_1,\ldots,s_Z]$, where $s_i\in\mathbb{R}^{H\times W}$ is a slice. A local encoder $f_l(\cdot;\theta_l)$ produces $z_{l,i}=f_l(s_i)\in\mathbb{R}^{d_l}$. A global encoder aggregates the ordered local tokens:
\begin{equation}
z_g=f_g(z_{l,1},\ldots,z_{l,Z};\theta_g)\in\mathbb{R}^{d_g}.
\end{equation}

Two stochastic, anatomy-preserving views are produced for each input. Both the local and global spaces use VICReg. For paired batches $Z$ and $Z'$, its objective is
\begin{equation}
\mathcal{L}_{\mathrm{VR}}(Z,Z')=\lambda s(Z,Z')+\mu[v(Z)+v(Z')]+\nu[c(Z)+c(Z')],
\end{equation}
where $s$, $v$, and $c$ are the invariance, variance, and covariance terms, respectively. The hyperparameters $\lambda$, $\mu$, and $\nu$ control the weight of view invariance, feature dispersion, and feature decorrelation. Let $Z_l^{(1)}$ and $Z_l^{(2)}$ denote the batches of projected local representations, and let $Z_g^{(1)}$ and $Z_g^{(2)}$ denote their projected global representation batches. The local and global objectives are
\begin{equation}
\mathcal{L}_l=\mathcal{L}_{\mathrm{VR}}\!\left(Z_l^{(1)},Z_l^{(2)}\right),\qquad
\mathcal{L}_g=\mathcal{L}_{\mathrm{VR}}\!\left(Z_g^{(1)},Z_g^{(2)}\right).
\end{equation}
The overall hierarchical pretraining objective is $\mathcal{L}=\mathcal{L}_l+\mathcal{L}_g$, where $\mathcal{L}_l$ optimizes intra-slice representations and $\mathcal{L}_g$ optimizes inter-slice volumetric representations.

More explicitly, let $A, B\in\mathbb{R}^{N\times D}$ denote the projected representations of two augmented views in one of the two spaces. The invariance term aligns paired samples,
\begin{equation}
s(A,B)=\frac{1}{N}\sum_{n=1}^{N}\lVert A_n-B_n\rVert_2^2.
\end{equation}
For feature dimension $j$, let $\sigma_j(A)=\sqrt{\operatorname{Var}(A_{:,j})+\epsilon}$. The variance term maintains dispersion above margin $\gamma$,
\begin{equation}
v(A)=\frac{1}{D}\sum_{j=1}^{D}\max(0,\gamma-\sigma_j(A)).
\end{equation}
After centering $A$ across its batch dimension, let $C(A)$ be its empirical feature covariance matrix. The covariance term reduces redundancy without forcing different samples apart,
\begin{equation}
c(A)=\frac{1}{D}\sum_{i\neq j}C(A)_{ij}^{2}.
\end{equation}
The three terms serve different purposes: invariance makes the representation stable to the permitted augmentations, variance prevents every input from mapping to a constant vector, and covariance discourages multiple coordinates from encoding the same variation. The local and global branches use the same mathematical form but compute their statistics over different representation spaces. Local batches contain projected slice embeddings; global batches contain projected volume embeddings. Consequently, the loss parameters are shared conceptually, not through a common projector or a mixed batch of local and global features.

To isolate local representation from the global objective, we apply stop-gradient to the local embeddings before global encoding. The embeddings therefore contribute to volume construction during the forward pass, but the global loss does not propagate gradients to or update the local encoder, thus separating optimization across the two spatial levels without implying statistical independence between their representations.

\begin{figure*}[t]
\centering
\includegraphics[width=0.92\textwidth]{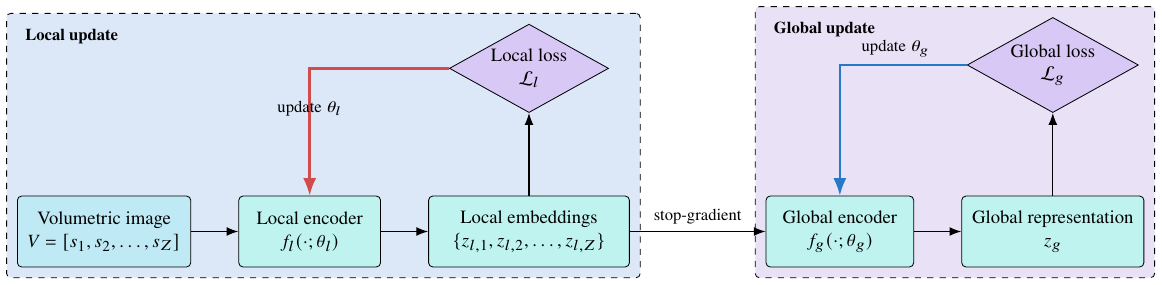}
\caption{Computational graph of L2GSSL. Local embeddings are inputs to the global encoder, but the global loss is detached at the local encoder. Solid arrows show forward computation; colored arrows show parameter updates.}
\label{fig:flow}
\end{figure*}

The asymmetric gradient flow in Fig.~\ref{fig:flow} separates the optimization of the two representation levels. The local objective encourages the slice encoder to preserve view-stable intra-slice content, whereas the global objective learns relationships among the ordered local embeddings. Applying stop-gradient before global encoding allows local embeddings to contribute to the volume representation while preventing the global objective from updating the local encoder. Thus, the two representation spaces remain functionally coupled during the forward pass but are optimized through separate gradient paths.

\subsection{Backbone Instantiations}
We instantiate L2GSSL with two backbone designs. In the Medical Slice Transformer (MST), a ViT image encoder extracts local tokens from each slice, and a slice transformer models inter-slice context. For the MST experiments, the local ViT encoder is initialized with publicly available DINOv2 weights~\cite{oquab2024dinov2} before L2GSSL pretraining on medical images. This architecture provides an explicit boundary between slice encoding and sequence aggregation. In the MaxViT instantiation, the backbone is initialized with ImageNet-pretrained weights, with the stem and first two stages forming the local encoder and the remaining two stages forming the global encoder.p This partition follows the backbone hierarchy rather than an explicit slice-aggregation boundary: early stages learn fine-grained spatial features, and later stages learn higher-level representations. The two instantiations test L2GSSL across an explicit slice transformer and a staged convolution-attention network.

\subsection{Slice-Wise Execution and Phased Optimization}
Slices are processed in mini-batches, and their embeddings are accumulated in anatomical order before global encoding. This execution pattern reduces the peak activation memory needed for stored local activations and enables volume modeling from compact tokens. Training follows three phases: local-only optimization for the first $30\%$ of epochs, joint local and global optimization for the next $40\%$, and global-only optimization for the final $30\%$. Stop-gradient prevents the global objective from updating the local encoder during the joint phase. In the final phase, the local encoder is frozen while the global encoder is optimized on its fixed embeddings. This schedule allows the global encoder to operate on local representations that have first undergone dedicated pretraining.

The phases control when each representation level becomes trainable. The local-only phase establishes slice features without requiring an initially untrained global encoder to provide a useful learning signal. During the joint phase, the local encoder continues to follow its own objective while the global encoder begins to model the ordered embeddings; the stop-gradient keeps these updates on separate parameter paths. The global-only phase then fixes the local representation distribution and concentrates the remaining optimization on inter-slice relationships. This schedule changes the optimization order rather than the downstream architecture, and the component study evaluates how its value changes with encoder initialization.

Let $b$ be the number of slices processed at once, and let $K=\lceil Z/b\rceil$ be the number of contiguous slice groups. For each volume, the paired augmented views are processed as follows:
\begin{enumerate}
\item Select the active objectives and trainable parameter groups for the current phase: $(\mathcal{L}_l,\theta_l)$ in the local-only phase, both $(\mathcal{L}_l,\theta_l)$ and $(\mathcal{L}_g,\theta_g)$ in the joint phase, and $(\mathcal{L}_g,\theta_g)$ in the global-only phase.
\item Partition the ordered slices into $K$ contiguous groups. When the local objective is active, forward one group at a time through the local encoder and projector, scale the group's local loss by $1/K$, and backpropagate it to accumulate parameter gradients. When the global objective is also active, append a detached copy of each group's embeddings in the original $z$-axis order.
\item In the global-only phase, keep the local encoder frozen and generate and store the ordered embeddings without constructing a local backward graph.
\item When the global objective is active, concatenate the stored embeddings at the stop-gradient boundary, forward the sequence through the global encoder and projector, and backpropagate $\mathcal{L}_g$.
\item Apply the optimizer steps for the parameter groups active in the current phase. After pretraining, discard the projector heads and pass the pretrained volume representation to the downstream classifier.
\end{enumerate}

Accordingly, the phase-specific execution paths are Steps 1, 2, and 5 for the local-only phase; Steps 1, 2, 4, and 5 for the joint phase; and Steps 1, 3, 4, and 5 for the global-only phase. Step 3 prepares the global input without local backpropagation, whereas Step 4 optimizes the global encoder and therefore applies to both the joint and global-only phases.

Processing slice groups sequentially avoids storing all slice activations simultaneously, and the global encoder operates on the output compact embedding sequence. Gradient accumulation approximates a larger optimization batch, but normalization statistics, when present, are computed within each slice group.

\subsection{Peak-Activation Analysis}
The activation-memory advantage can be expressed independently of a particular GPU. Let $a_l$ denote the storage required to retain the local-encoder computation graph for one slice, let $a_l^{\mathrm{fwd}}$ denote the temporary storage for a frozen local-encoder forward pass, let $e_l$ denote the storage for one detached local embedding, and let $A_g(Z)$ denote the internal activation storage of the global encoder for a sequence of $Z$ embeddings. Parameters, optimizer state, input, and framework-allocation memory are omitted because this analysis isolates the activation terms affected by slice grouping and gradient detachment.

A conventional end-to-end eager implementation without activation recomputation or offloading retains the local computation graphs for all slices until the global backward pass, giving
\begin{equation}
M_{\mathrm{e2e}}=O\!\left(Za_l+A_g(Z)\right).
\end{equation}
During the L2GSSL joint phase, each local group completes its backward pass before global encoding. The detached embeddings persist across groups, but the local and global activation peaks occur at different times. The corresponding peak storage is therefore
\begin{equation}
M_{\mathrm{L2GSSL}}^{\mathrm{joint}}=O\!\left(Ze_l+\max\{ba_l,A_g(Z)\}\right),\qquad b\ll Z.
\end{equation}
The local-only phase requires $O(ba_l)$ activation storage because no global sequence is constructed. In the global-only phase, the local encoder is frozen, giving $O\!\left(Ze_l+\max\{ba_l^{\mathrm{fwd}}, A_g(Z)\}\right)$. The reduction applies to the local-activation term; the ordered embedding sequence and global encoder still scale with the number of slices. Slice grouping also does not reduce the number of local-encoder forward operations required for a volume, and groups trained with the local objective still require backward computation. Sequential groups may introduce additional execution overhead. The analysis therefore explains why the design can lower peak activation storage under the stated execution model, but it does not imply lower wall-clock time or higher data-ingestion throughput. The group size $b$ exposes the practical trade-off: smaller groups reduce the local-activation term, whereas larger groups may improve within-group hardware utilization when memory permits.

%% file: content/exp.tex
\section{Experiments}
\subsection{Datasets and Tasks}

LIDC-IDRI contains $1{,}018$ thoracic CT cases with multi-reader lung-nodule annotations~\cite{armato2011lidc}. After nodule-level processing, the study contains $1{,}625$ samples; malignancy tendency is formulated as a binary classification task. Splitting is performed at the patient/scan level before crop or slice extraction, ensuring that related samples from a scan remain in one partition. Fine-tuning uses $10\%$, $25\%$, or $100\%$ of the training labels, corresponding to $104$, $260$, and $1{,}040$ nodules.

LNDb contains $294$ chest CT scans from a Portuguese institution and annotations from five participating radiologists~\cite{pedrosa2019lndb}. We retain nodules with malignancy assessments and align the label definition and preprocessing with LIDC-IDRI. Models fine-tuned using $25\%$ of LIDC-IDRI are evaluated directly on LNDb; no LNDb fine-tuning is performed.

For task-level breadth, the RSNA abdominal-trauma data contain $3{,}147$ patients collected from $23$ institutions in $14$ countries~\cite{hermans2025rsna}. The task predicts kidney, liver, spleen, bowel, and extravasation findings. AbdomenCT-1K supplies additional unlabeled abdominal CT for pretraining~\cite{ma2021abdomenct}. We integrate L2GSSL initialization into four experts of the second-place solution to the 2023 RSNA Abdominal Trauma Detection competition while preserving its downstream procedures.

Table~\ref{tab:datasets} separates the role of each collection. LIDC-IDRI is the primary controlled benchmark because it supports nodule-level low-label experiments and multi-reader analysis. LNDb is held outside training and model adaptation to test direct cross-dataset transfer. RSNA and AbdomenCT-1K form a second anatomical setting: the former supplies the downstream abdominal-trauma task, while the latter contributes anatomically aligned volumes during pretraining. This design tests label efficiency, source transfer, and task breadth without treating all datasets as interchangeable or combining their labels.

\begin{table*}[t]
\centering
\caption{Datasets and their roles in the evaluation.}
\label{tab:datasets}
\resizebox{\textwidth}{!}{\begin{tabular}{llllll}
\toprule
Dataset & Anatomy/modality & Available scale & Use in this study & Label use & Evaluation role\\
\midrule
LIDC-IDRI~\cite{armato2011lidc} & Thoracic CT & $1{,}018$ cases; $1{,}625$ processed nodules & SSL and downstream training & $104$, $260$, or $1{,}040$ training nodules & Internal discrimination and reader analysis\\
LNDb~\cite{pedrosa2019lndb} & Thoracic CT & $294$ scans & External testing only & No target-domain training & Direct cross-dataset transfer\\
RSNA 2023~\cite{hermans2025rsna} & Abdominal CT & $3{,}147$ patients & Downstream trauma pipeline & Competition task labels & Multi-organ task transfer\\
AbdomenCT-1K~\cite{ma2021abdomenct} & Abdominal CT & Auxiliary volumetric collection & Additional SSL source & Labels not used by SSL & Anatomically aligned pretraining\\
\bottomrule
\end{tabular}}
\end{table*}

The LIDC-IDRI split is created before any nodule crop or slice is extracted. This order is essential: splitting the derived samples allows visually correlated slices or nodules from one scan to enter both training and testing. All compared methods use the same leakage-controlled partition. LNDb is processed with the LIDC-IDRI pipeline and the matching malignancy definition, but ambiguous cases without a usable malignancy assessment are not converted into definite labels. No LNDb sample contributes to pretraining, fine-tuning, or checkpoint selection in the reported external evaluation.

\subsection{Experimental Setup and Evaluation Protocols}
We compare supervised MST-DINOv2, MoCo v2, SimSiam, volume-level VICReg, and L2GSSL using the same backbone and pipeline where applicable.

For downstream MST evaluation, linear probing trains only the classifier, whereas full fine-tuning updates the entire network. All compared methods use the same downstream optimization and test-time augmentation protocol, and checkpoints are selected using the validation area under the receiver operating characteristic curve. Linear-probing results use three seeds, and fine-tuning uses five runs. Complete hyperparameter settings are provided in the released code.

CT values are clipped to $[-1000,1000]$ Hounsfield units and scaled to $[-1,1]$. Volumes are resampled as needed, and crop-or-pad uses the minimum volume intensity for padding. SSL views use random gamma correction (probability $0.3$, log-gamma $[-0.2,0.2]$), low-frequency intensity distortion (probability $0.2$, coefficient magnitude $0.5$), Gaussian blur (probability $0.2$, standard deviation $[0,1.0]$), affine transformation (probability $0.5$, scale $[0.95,1.05]$, rotation up to $10^\circ$), and Gaussian noise (probability $0.3$, standard deviation $[0,0.05]$).

We report the area under the receiver operating characteristic curve (AUROC) and the area under the precision--recall curve (AUPRC) for discrimination under class imbalance. Means and standard deviations summarize independent runs.

The protocols address distinct evaluation questions rather than form a single cross-dataset ranking. LIDC-IDRI supports controlled representation and label-efficiency comparisons, LNDb evaluates direct transfer without target-domain adaptation, and the RSNA/AbdomenCT-1K setting evaluates transfer across anatomy and downstream task. Within LIDC-IDRI, the protocols distinguish fixed-batch and memory-capped representation evaluation from fine-tuning at different label fractions. Table~\ref{tab:protocols} summarizes the controlled factor and primary question for each protocol.

\begin{table*}[t]
\centering
\caption{Evaluation protocols and the claims they are designed to test.}
\label{tab:protocols}
\resizebox{\textwidth}{!}{\begin{tabular}{lllll}
\toprule
Protocol & Encoder update & Supervision & Controlled factor & Primary question\\
\midrule
Fixed-batch linear probe & Frozen & LIDC-IDRI labels & Local batch $8$; two accumulation steps & Are pretrained features linearly separable?\\
VRAM-capped linear probe & Frozen & LIDC-IDRI labels & Peak VRAM constrained to $47$--$48$ GB & How effectively is a fixed memory budget used?\\
Label-fraction fine-tuning & Full network & $10\%$, $25\%$, or $100\%$ of LIDC-IDRI & Shared split and downstream pipeline & Does pretraining reduce dependence on labels?\\
Direct LNDb evaluation & Frozen at inference & No LNDb training labels & LIDC-IDRI model evaluated without adaptation & Does performance transfer across data sources?\\
RSNA pipeline integration & Full downstream pipeline & RSNA task labels & Original multi-expert inference design retained & Does initialization transfer to abdominal trauma CT?\\
\bottomrule
\end{tabular}}
\end{table*}

\subsection{Label-Efficient Transfer on LIDC-IDRI}
We evaluate representation separability and label efficiency under controlled LIDC-IDRI protocols.


Table~\ref{tab:finetune} shows that L2GSSL is most beneficial when annotations are limited. With $10\%$ and $25\%$ of labels, it achieves the highest AUROC and AUPRC among the evaluated methods. Relative to volume-level VICReg, the AUROC gains are $0.0229$ and $0.0294$, respectively. When all training labels are used, supervised MST-DINOv2 has the highest AUROC by $0.0044$, while SimSiam has the highest AUPRC by $0.0080$. The diminishing margin is consistent with SSL serving primarily as an initialization advantage when labels are scarce.

\begin{table*}[t]
\centering
\caption{LIDC-IDRI full fine-tuning under identical batch-size constraints. Best values are bold.}
\label{tab:finetune}
\resizebox{\textwidth}{!}{\begin{tabular}{lcccccc}
\toprule
& \multicolumn{2}{c}{$10\%$ labels ($104$)} & \multicolumn{2}{c}{$25\%$ labels ($260$)} & \multicolumn{2}{c}{$100\%$ labels ($1{,}040$)}\\
Method & AUROC & AUPRC & AUROC & AUPRC & AUROC & AUPRC\\
\midrule
Supervised MST-DINOv2 & $0.6564\pm0.0789$ & $0.4557\pm0.0706$ & $0.8695\pm0.0499$ & $0.7875\pm0.0781$ & $\mathbf{0.9448}\pm0.0070$ & $0.9044\pm0.0070$\\
MoCo v2 & $0.5986\pm0.0175$ & $0.4240\pm0.0190$ & $0.8221\pm0.0727$ & $0.7290\pm0.1276$ & $0.9360\pm0.0073$ & $0.8893\pm0.0128$\\
SimSiam & $0.6847\pm0.1026$ & $0.5033\pm0.1212$ & $0.8007\pm0.1462$ & $0.6936\pm0.1966$ & $0.9389\pm0.0024$ & $\mathbf{0.9123}\pm0.0055$\\
Volume VICReg & $0.9000\pm0.0048$ & $0.8573\pm0.0034$ & $0.9083\pm0.0022$ & $0.8686\pm0.0133$ & $0.9296\pm0.0077$ & $0.8867\pm0.0092$\\
L2GSSL & $\mathbf{0.9229}\pm0.0100$ & $\mathbf{0.8769}\pm0.0042$ & $\mathbf{0.9377}\pm0.0039$ & $\mathbf{0.9003}\pm0.0064$ & $0.9404\pm0.0024$ & $0.9043\pm0.0038$\\
\bottomrule
\end{tabular}}
\end{table*}

Linear probing isolates representation quality. Under identical batch size (Table~\ref{tab:linear}), L2GSSL improves over volume-level VICReg in both AUROC and AUPRC.

\begin{table}[t]
\centering
\caption{LIDC-IDRI linear probing with identical batch size.}
\label{tab:linear}
\begin{tabular}{lcc}
\toprule
Method & AUROC & AUPRC\\
\midrule
MoCo v2 & $.6749\pm.0039$ & $.5086\pm.0012$\\
SimSiam & $.6226\pm.0092$ & $.4545\pm.0167$\\
Volume VICReg & $.8485\pm.0092$ & $.7740\pm.0109$\\
L2GSSL & $\mathbf{.8875}\pm.0061$ & $\mathbf{.8111}\pm.0118$\\
\bottomrule
\end{tabular}
\end{table}

\subsection{Cross-Dataset and Task Transfer}
\subsubsection{Direct Cross-Dataset Evaluation}
We evaluate whether representations learned on LIDC-IDRI transfer to a different data source without target-domain adaptation.

Table~\ref{tab:lndb} reports direct LNDb evaluation. L2GSSL is best in both discrimination metrics. Compared with volume-level VICReg, the strongest evaluated SSL baseline, it gains $0.1130$ AUROC and $0.0954$ AUPRC. Because no target-domain adaptation is performed, this experiment provides stronger evidence of transfer than another internal split.

\begin{table}[t]
\centering
\caption{Direct LNDb evaluation after fine-tuning on $25\%$ of LIDC-IDRI; no LNDb adaptation.}
\label{tab:lndb}
\begin{tabular}{lcc}
\toprule
Method & AUROC & AUPRC\\
\midrule
Supervised & $.6869\pm.0553$ & $.6289\pm.0632$\\
MoCo v2 & $.6717\pm.0245$ & $.6215\pm.0075$\\
SimSiam & $.6313\pm.1005$ & $.5551\pm.1603$\\
Volume VICReg & $.7008\pm.0154$ & $.6635\pm.0515$\\
L2GSSL & $\mathbf{.8138}\pm.0401$ & $\mathbf{.7589}\pm.0461$\\
\bottomrule
\end{tabular}
\end{table}

\subsubsection{Abdominal Trauma Transfer}
We evaluate whether L2GSSL initialization transfers beyond lung-nodule classification to abdominal CT and a multi-organ trauma-prediction pipeline.

For abdominal-trauma pretraining, the projector is a three-layer multilayer perceptron with hidden dimension $4096$, and the evaluation uses severity-weighted RSNA competition loss. In the RSNA multi-expert pipeline, the original competition loss is $0.3597$, while the best configuration incorporating L2GSSL-pretrained experts obtains $0.3551$. This result shows that L2GSSL can use additional unlabeled anatomically aligned AbdomenCT-1K volumes without changing the downstream inference design.

\subsection{Resource and Volumetric Baseline Comparisons}
\subsubsection{Common GPU-Memory Budget}
The VRAM-capped experiment complements the fixed-batch comparison by replacing an identical optimization configuration with a common GPU-memory envelope for the MST-based SSL methods. Each SSL method is constrained to a peak VRAM budget of $47$--$48$ GB, reflecting the practical case in which the accelerator capacity is fixed. This comparison evaluates the representation quality achievable within the same memory envelope rather than isolating the SSL objective alone. Table~\ref{tab:memory-cap} shows that L2GSSL achieves the highest AUROC and AUPRC under the common cap, followed by volume-level VICReg.

\begin{table}[t]
\centering
\caption{LIDC-IDRI linear probing with the MST-based methods constrained to a common peak VRAM budget. Best results are bold.}
\label{tab:memory-cap}
\resizebox{\columnwidth}{!}{\begin{tabular}{lccc}
\toprule
Method & Peak VRAM (GB) & AUROC & AUPRC\\
\midrule
MoCo v2 & $47$--$48$ & $.6661\pm.0104$ & $.4936\pm.0037$\\
SimSiam & $47$--$48$ & $.6226\pm.0092$ & $.4545\pm.0167$\\
Volume VICReg & $47$--$48$ & $.8485\pm.0092$ & $.7740\pm.0109$\\
L2GSSL & $47$--$48$ & $\mathbf{.8599}\pm.0038$ & $\mathbf{.7926}\pm.0039$\\
\bottomrule
\end{tabular}}
\end{table}

\subsubsection{Fully 3D Reference}
VoCo operates directly on dense 3D crops and does not share the slice-encoder and inter-slice aggregation structure used by the MST-based methods. Incorporating it into the controlled comparisons above would therefore require changing its backbone, input formulation, and training procedure. We instead evaluate VoCo separately as a fully 3D system-level reference. We adapt the extended VoCo formulation~\cite{wu2024largescale} to the LIDC-IDRI nodule setting using a 3D Swin encoder and VoCo-style volumetric crops, and evaluate both random and MONAI-pretrained initialization. Complete preprocessing and optimization settings are provided in the released code. Table~\ref{tab:voco} shows that L2GSSL uses slightly less measured peak VRAM and achieves higher linear probing and fine-tuning performance than both VoCo variants in all evaluated settings: linear probing and fine-tuning under $10\%$, $25\%$, and $100\%$ available labels.

\begin{table*}[t]
\centering
\caption{Comparison with the fully 3D VoCo reference on LIDC-IDRI. Linear probing (LP), fine-tuning (FT), and measured peak VRAM are reported for the evaluated configurations. The comparison is not controlled for backbone, input formulation, or initialization. Best discrimination results are bold.}
\label{tab:voco}
\resizebox{\textwidth}{!}{\begin{tabular}{lccccccccc}
\toprule
& & \multicolumn{2}{c}{LP} & \multicolumn{2}{c}{FT, $10\%$ labels} & \multicolumn{2}{c}{FT, $25\%$ labels} & \multicolumn{2}{c}{FT, $100\%$ labels}\\
Method & Peak VRAM (GB) & AUROC & AUPRC & AUROC & AUPRC & AUROC & AUPRC & AUROC & AUPRC\\
\midrule
VoCo (scratch) & $27.57$ & $.6770\pm.0085$ & $.4775\pm.0193$ & $.6324\pm.0669$ & $.4523\pm.0734$ & $.7001\pm.0290$ & $.5286\pm.0436$ & $.7805\pm.0166$ & $.6482\pm.0010$\\
VoCo (MONAI) & $27.57$ & $.6905\pm.0065$ & $.5361\pm.0069$ & $.6508\pm.0561$ & $.4641\pm.0512$ & $.7058\pm.0231$ & $.5491\pm.0267$ & $.7654\pm.0118$ & $.6182\pm.0188$\\
L2GSSL & $24.37$--$24.48$ & $\mathbf{.8875}\pm.0061$ & $\mathbf{.8111}\pm.0118$ & $\mathbf{.9229}\pm.0100$ & $\mathbf{.8769}\pm.0042$ & $\mathbf{.9377}\pm.0039$ & $\mathbf{.9003}\pm.0064$ & $\mathbf{.9404}\pm.0024$ & $\mathbf{.9043}\pm.0038$\\
\bottomrule
\end{tabular}}
\end{table*}

\subsection{Backbone Portability}
We test whether the local-to-global formulation remains effective when instantiated with a second backbone.

Table~\ref{tab:maxvit} shows that L2GSSL achieves the highest AUROC and AUPRC with MaxViT. These results support portability beyond MST without implying universality across architectures.

\begin{table}[t]
\centering
\caption{LIDC-IDRI linear probing with the MaxViT backbone. Best results are bold.}
\label{tab:maxvit}
\begin{tabular}{lcc}
\toprule
Method & AUROC & AUPRC\\
\midrule
MoCo v2 & $.6999\pm.0513$ & $.6842\pm.0357$\\
SimSiam & $.7510\pm.0054$ & $.7253\pm.0057$\\
Volume VICReg & $.7567\pm.0562$ & $.7400\pm.0542$\\
L2GSSL & $\mathbf{.8919}\pm.0051$ & $\mathbf{.8850}\pm.0034$\\
\bottomrule
\end{tabular}
\end{table}

\subsection{Model-Behavior Analyses}
\subsubsection{Lesion Dependence}
We test whether the predictions depend on lesion-bearing content by masking the corresponding slices at inference. Table~\ref{tab:mask} shows a $73.27$ percentage-point decline in malignant true-positive rate, together with large AUROC and AUPRC drops. The result indicates that the model relies substantially on information in malignant lesion slices rather than only on background or dataset-level shortcuts.

\begin{table}[t]
\centering
\caption{Effect of masking malignant lesion-bearing slices.}
\label{tab:mask}
\begin{tabular}{lccc}
\toprule
Condition & Malignant TPR & AUROC & AUPRC\\
\midrule
No mask & $87.84\%$ & $0.9404$ & $0.9043$\\
Tumor-slice mask & $14.57\%$ & $0.6169$ & $0.4307$\\
Change & $-73.27\%$ & $-0.3235$ & $-0.4736$\\
\bottomrule
\end{tabular}
\end{table}

\subsubsection{Radiologist Context and Sensitivity to Uncertain Cases}
This analysis serves two purposes. First, it places the model ROC curves in the context of a pooled operating point derived from per-nodule held-out radiologist judgments. Second, it assesses how model discrimination changes when the ROC evaluation retains less-certain cases in which the held-out radiologist assigns a score of $3$ despite a definite reference from the remaining radiologists.

We use LIDC-IDRI for this comparison because it provides multiple radiologist malignancy ratings for individual nodules, which are required to construct a reference that excludes the radiologist being evaluated. LNDb does not provide comparable multi-reader coverage for each nodule. 

For each nodule, one of its annotating radiologists is randomly selected as the held-out radiologist. Selection is performed per nodule because radiologist coverage differs across the test set and no single radiologist annotates every included nodule. The mean score of the remaining radiologists defines the reference label: means at or below $2.5$ are benign, means at or above $3.5$ are malignant, and intermediate means are excluded. We then define two evaluation conditions according to how we handle an uncertain score of $3$ from the held-out radiologist. The uncertainty-excluded condition omits cases in which the held-out radiologist assigns a score of $3$ and therefore evaluates only definite held-out judgments. The uncertainty-penalized condition retains such cases and counts the uncertain judgment as disagreement whenever the remaining radiologists provide a definite reference. After these rules are applied, the held-out judgments from all included nodules are pooled to compute a single sensitivity and false-positive rate rather than averaging radiologist-specific metrics. This design prevents the selected radiologist from contributing to the label against which that radiologist is compared.

The pooled radiologist point provides an operating reference in the same ROC space as the model curves by summarizing per-nodule held-out judgments across the evaluation set. The uncertainty-penalized condition is expected by construction to reduce the pooled radiologist point because a score of $3$ is counted as disagreement against a definite reference. The more informative observation is that the L2GSSL ROC curves shift only modestly when these cases are retained in the ROC evaluation and remain close across runs. The analysis therefore provides performance context together with a sensitivity check on reader-uncertain cases.

\begin{figure*}[t]
\centering
\subfloat[Uncertainty-excluded scoring.]{\includegraphics[width=0.48\textwidth]{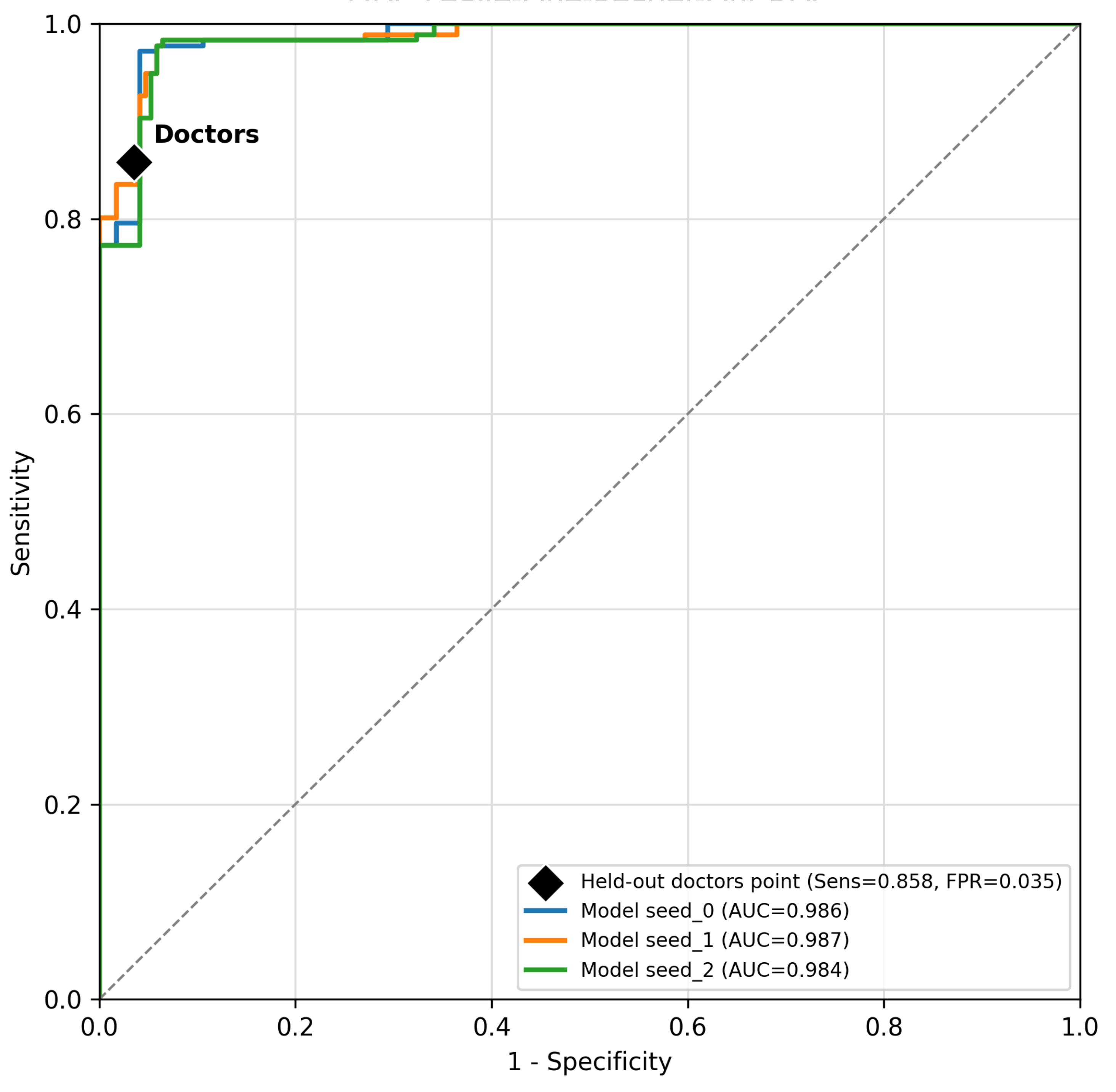}\label{fig:reader-excluded}}
\hfill
\subfloat[Uncertainty-penalized scoring.]{\includegraphics[width=0.48\textwidth]{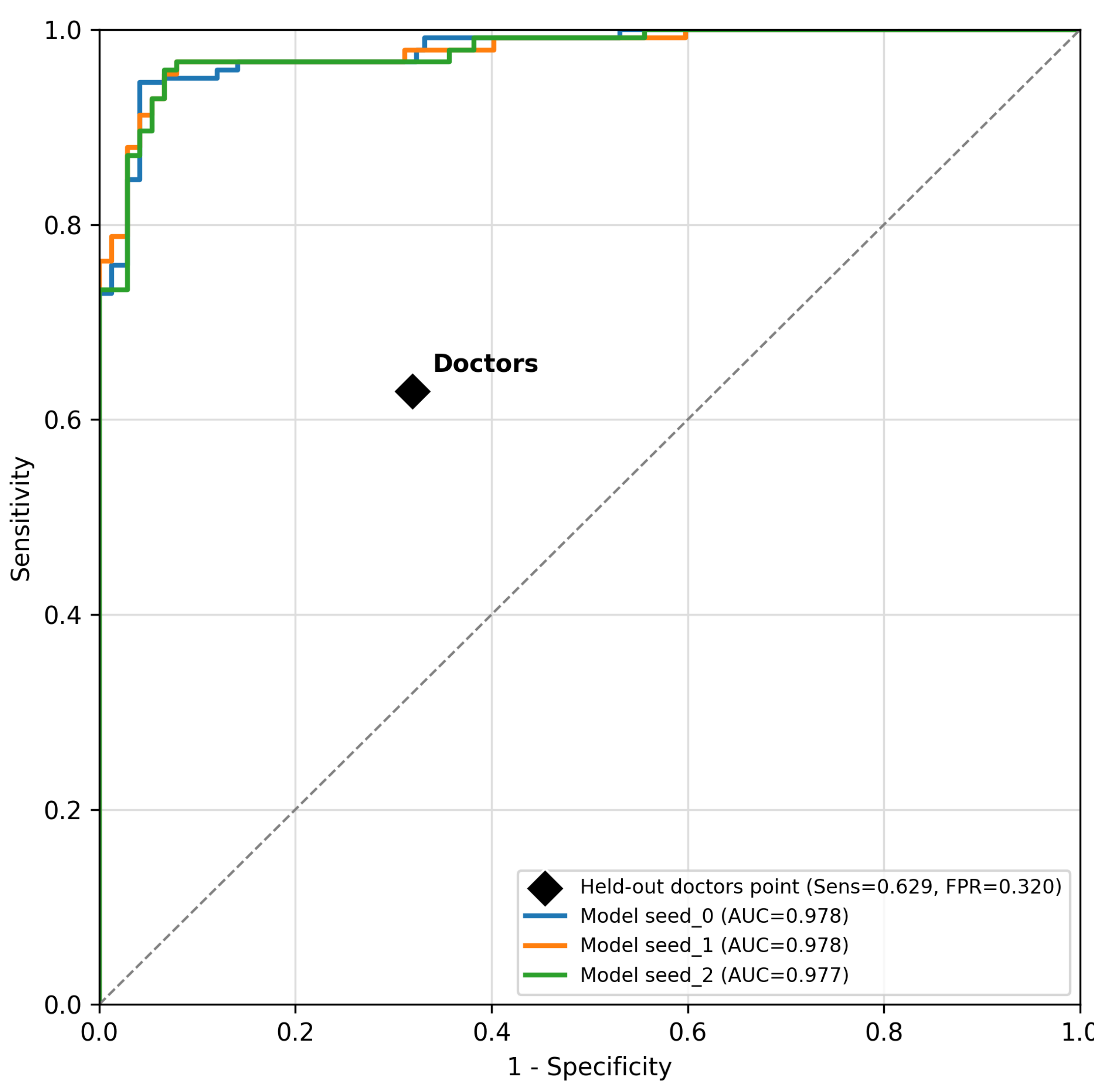}\label{fig:reader-penalized}}
\caption{Per-nodule held-out-radiologist ROC analysis under two treatments of an uncertain score from the selected radiologist. Each reference label is constructed from the remaining radiologists. The diamond shows the sensitivity and false-positive rate computed after pooling the held-out judgments across nodules; curves show L2GSSL runs.}
\label{fig:reader}
\end{figure*}

\subsection{Design and Implementation Sensitivity}
\subsubsection{Component Ablation}
We examine how the stop-gradient boundary and phased optimization contribute to L2GSSL behavior.

Removing the stop-gradient boundary gives the weakest discrimination results under both initialization strategies, supporting the role of gradient isolation in local feature learning (Table~\ref{tab:component-ablation}). The effect of phased optimization depends on initialization: the full configuration performs best from random initialization, whereas removing the phased schedule performs best from DINOv2 initialization. The schedule therefore acts as an initialization-dependent optimization mechanism rather than a universally beneficial component.

\begin{table}[t]
\centering
\caption{Linear-probing component ablation on LIDC-IDRI. Results are mean $\pm$ standard deviation; the best result within each initialization is bold.}
\label{tab:component-ablation}
\setlength{\tabcolsep}{3.5pt}
\resizebox{\columnwidth}{!}{%
\begin{tabular}{llcc}
\toprule
Initialization & Configuration & AUROC & AUPRC\\
\midrule
Random & Full L2GSSL & \textbf{$0.5968\pm0.0011$} & \textbf{$0.3938\pm0.0026$}\\
Random & Without stop-gradient & $0.5767\pm0.0202$ & $0.3757\pm0.0054$\\
Random & Without phased schedule & $0.5877\pm0.0191$ & $0.3801\pm0.0183$\\
\midrule
DINOv2 & Full L2GSSL & $0.8599\pm0.0038$ & $0.7926\pm0.0039$\\
DINOv2 & Without stop-gradient & $0.5749\pm0.0299$ & $0.3806\pm0.0207$\\
DINOv2 & Without phased schedule & \textbf{$0.8967\pm0.0028$} & \textbf{$0.8260\pm0.0060$}\\
\bottomrule
\end{tabular}}
\end{table}

\subsubsection{Projector Normalization Sensitivity}
A separate sensitivity analysis examines whether the weaker MoCo v2 and SimSiam results arise from the absence of batch normalization in their projectors. They do not: with projector batch normalization, MoCo v2 records $0.6497\pm0.0014$ AUROC and $0.4889\pm0.0357$ AUPRC, SimSiam records $0.5731\pm0.0008$ and $0.3195\pm0.0009$, and L2GSSL records $0.7623\pm0.0071$ and $0.5615\pm0.0045$. In slice-wise accumulation, normalization statistics are estimated within each micro-batch even though gradients approximate a larger effective batch. The mismatch produces noisy statistics and can degrade training; the main comparisons therefore retain the original normalization setting.

%% file: content/disc.tex
\section{Discussion}
\subsection{Principal Findings}
The results show that separating slice- and volume-level self-supervised objectives improves label-efficient learning from volumetric CT data. L2GSSL provides its clearest gains when only $10\%$ or $25\%$ of the LIDC-IDRI labels are available, while the advantage narrows under full-label fine-tuning. This pattern is consistent with the intended role of self-supervised pretraining: using unlabeled scans to reduce dependence on curated annotations. Direct evaluation on LNDb further tests whether the learned representation remains useful under a change in data source and annotation practice, while the abdominal-trauma task extends the evaluation across anatomy and prediction target.

One interpretation of the label-fraction trend follows from the two representation levels. The local objective supplies view-stable slice features before downstream labels are introduced, while the global objective organizes those features across the ordered volume. With fewer labels, this initialization reduces how much slice appearance and inter-slice structure must be learned from supervised examples alone. Full-label fine-tuning supplies more task-specific supervision to all trainable components and can therefore narrow differences among initializations. This interpretation is consistent with the observed trend and with the component analysis of gradient isolation and scheduling.

The resource experiments address a separate practical consideration: whether hierarchical volumetric learning remains feasible under constrained GPU memory. The evaluated results show that slice-to-volume learning can provide useful representations within the measured memory budgets. This property is relevant to volumetric healthcare archives, where scans are high-dimensional and vary in slice count even when labels remain limited.

The component results clarify how the optimization design contributes to this behavior. Under DINOv2 initialization, removing gradient detachment produces the weakest variant, whereas removing phased optimization can improve performance. With random initialization, the full phased configuration performs best, indicating that the schedule is most useful before a stable slice representation has been established. The ablation therefore supports controlled local-to-global optimization while showing that the preferred schedule depends on initialization.

\subsection{Implications for Healthcare Data Processing}
The intended use of L2GSSL is representation pretraining, not autonomous diagnosis. Within that scope, its modularity offers three practical advantages. First, unlabeled scans can contribute to adaptation before a site invests in outcome curation. This matters because routine archives are far larger than the subsets with consistent malignancy, injury, or segmentation labels. Second, slice-wise execution decouples storage and optimization choices: images can be decoded in groups, local features can be accumulated, and the inter-slice model can operate on a sequence of embeddings. This design fits healthcare archives in which complete volumes vary in slice count and cannot all be staged as dense tensors at once. Third, the method retains a recognizable boundary between image appearance and volume context, which provides intervention points for auditing local features, sequence aggregation, and final predictions separately.

The framework does not eliminate the governance requirements associated with such archives. Site-specific secondary pretraining could absorb scanner signatures, reconstruction kernels, or population correlates that are unrelated to pathology. De-identification does not prevent these shortcuts, and SSL provides no label-based mechanism for detecting them during pretraining. A deployment pipeline should therefore retain provenance for every pretrained scan, document inclusion and exclusion rules, stratify evaluation by acquisition source where metadata permit, and keep test institutions outside all pretraining and hyperparameter selection. The patient-level split used here prevents within-dataset leakage, and the untouched LNDb test provides an initial distribution-shift check, but neither substitutes for institutional governance.

\subsection{Clinical Interpretation of the Evaluation}
The low-label experiment approximates a development setting with few curated nodules, but a label fraction is not equivalent to a clinical annotation budget. The selected samples may not capture the spectrum, prevalence, or disagreement encountered during prospective collection. Similarly, AUROC measures ranking across thresholds, whereas an actual decision-support tool operates at a chosen threshold with asymmetric consequences. AUPRC is included because malignant nodules are the positive class and class imbalance can make ROC performance appear optimistic, but neither measure specifies whether missed malignancies or false alarms are acceptable. Decision-curve analysis and task-specific utility were not performed.

The radiologist analysis provides a pooled operating reference for contextualizing the model ROC curves and tests whether model ranking remains stable when reader-uncertain cases are retained in the ROC evaluation. It is a retrospective analysis of existing per-nodule judgments; the radiologists did not interact with the model. Likewise, the LNDb experiment evaluates technical generalization rather than clinical transportability. A prospective study would need prespecified endpoints, locked preprocessing and thresholds, workflow timing, failure handling, and analysis of whether assistance changes radiologist decisions. These distinctions are important for a healthcare-data venue because a model can transfer statistically without improving care.

\subsection{Limitations}
The primary MST comparisons use the same patient-level split, backbone, and training pipeline to isolate differences among the evaluated pretraining objectives. VoCo, however, operates directly on dense 3D crops and cannot be mapped onto the slice-encoder and inter-slice aggregation structure without changing its original formulation. We therefore retain VoCo as a separate fully 3D system-level reference under the evaluated resource settings. Because VoCo and L2GSSL differ in backbone, input formulation, and initialization, this comparison evaluates the complete systems rather than isolating the effect of either objective.

The evaluation is limited to retrospective CT classification. LIDC-IDRI and LNDb provide variation in data source and reader protocol, and the RSNA task extends the evaluation to abdominal anatomy and multiple institutions, but these datasets do not cover prospective deployment, other imaging modalities, or demographic subgroup performance. The resource analysis measures peak VRAM rather than preprocessing throughput, storage, or training time. Broader claims would require evaluation on additional tasks, modalities, institutions, and prospective clinical workflows.

The lesion-masking experiment specifically examines whether model performance depends on lesion-bearing slices. The observed performance drop provides evidence that these slices contribute to the prediction, but their removal also changes the input distribution; the experiment therefore does not establish causal localization or exclude reliance on other image features.

%% file: content/conc.tex
\section{Conclusion}
L2GSSL addresses a central healthcare-data imbalance: high-dimensional CT volumes are widely available, while expert-curated labels remain limited. By decomposing self-supervised volumetric learning into local slice and global inter-slice objectives, the framework extracts reusable representations from unlabeled scans with controlled gradient flow and phased optimization. Its clearest gains occur with $10\%$ and $25\%$ of LIDC-IDRI labels and diminish under full supervision, matching the intended role of self-supervised pretraining. Direct LNDb transfer, two backbone designs, and abdominal-trauma CT provide evidence across data sources, architectures, anatomy, and tasks, while slice-wise execution supports training under constrained GPU memory. Together, these results position L2GSSL as a practical foundation for label-efficient and resource-conscious representation learning from volumetric healthcare data.

%% file: ref.bib
@article{chen2019context,
  author={Chen, Liang and Bentley, Paul and Mori, Kensaku and Misawa, Kazunari and Fujiwara, Michitaka and Rueckert, Daniel},
  title={Self-supervised learning for medical image analysis using image context restoration},
  journal={Medical Image Analysis},
  volume={58},
  pages={101539},
  year={2019},
  doi={10.1016/j.media.2019.101539}
}

@article{zhou2021genesis,
  author={Zhou, Zongwei and Sodha, Vatsal and Pang, Jiaxuan and Gotway, Michael B. and Liang, Jianming},
  title={Models {Genesis}},
  journal={Medical Image Analysis},
  volume={67},
  pages={101840},
  year={2021},
  doi={10.1016/j.media.2020.101840}
}

@inproceedings{wu2024voco,
  author={Wu, Linshan and Zhuang, Jiaxin and Chen, Hao},
  title={VoCo: A simple-yet-effective volume contrastive learning framework for 3D medical image analysis},
  booktitle={Proceedings of the IEEE/CVF Conference on Computer Vision and Pattern Recognition},
  pages={22873--22882},
  year={2024}
}

@article{wu2024largescale,
  author={Wu, Linshan and Zhuang, Jiaxin and Chen, Hao},
  title={Large-scale {3D} medical image pre-training with geometric context priors},
  journal={arXiv preprint arXiv:2410.09890},
  year={2024}
}

@inproceedings{bardes2022vicreg,
  author={Bardes, Adrien and Ponce, Jean and LeCun, Yann},
  title={VICReg: Variance-invariance-covariance regularization for self-supervised learning},
  booktitle={International Conference on Learning Representations},
  year={2022}
}

@article{armato2011lidc,
  author={Armato, III, Samuel G. and others},
  title={The Lung Image Database Consortium ({LIDC}) and Image Database Resource Initiative ({IDRI}): A completed reference database of lung nodules on {CT} scans},
  journal={Medical Physics},
  volume={38},
  number={2},
  pages={915--931},
  year={2011},
  doi={10.1118/1.3528204}
}

@article{pedrosa2019lndb,
  author={Pedrosa, Jo{\~a}o and others},
  title={{LNDb}: A lung nodule database on computed tomography},
  journal={arXiv preprint arXiv:1911.08434},
  year={2019}
}

@article{hermans2025rsna,
  author={Hermans, Sebastiaan and others},
  title={{RSNA} 2023 Abdominal Trauma {AI} Challenge: Review and outcomes},
  journal={Radiology: Artificial Intelligence},
  volume={7},
  number={1},
  pages={e240334},
  year={2025},
  doi={10.1148/ryai.240334}
}

@article{ma2021abdomenct,
  author={Ma, Jun and others},
  title={{AbdomenCT-1K}: Is abdominal organ segmentation a solved problem?},
  journal={IEEE Transactions on Pattern Analysis and Machine Intelligence},
  volume={44},
  number={10},
  pages={6695--6714},
  year={2022},
  doi={10.1109/TPAMI.2021.3100536}
}

@inproceedings{cicek20163d,
  author={{\c{C}}i{\c{c}}ek, {\"O}zg{\"u}n and Abdulkadir, Ahmed and Lienkamp, Soeren S. and Brox, Thomas and Ronneberger, Olaf},
  title={3D U-Net: Learning dense volumetric segmentation from sparse annotation},
  booktitle={Medical Image Computing and Computer-Assisted Intervention},
  pages={424--432},
  year={2016},
  doi={10.1007/978-3-319-46723-8_49}
}

@inproceedings{milletari2016vnet,
  author={Milletari, Fausto and Navab, Nassir and Ahmadi, Seyed-Ahmad},
  title={V-Net: Fully convolutional neural networks for volumetric medical image segmentation},
  booktitle={Fourth International Conference on 3D Vision},
  pages={565--571},
  year={2016},
  doi={10.1109/3DV.2016.79}
}

@inproceedings{hatamizadeh2022unetr,
  author={Hatamizadeh, Ali and Tang, Yucheng and Nath, Vishwesh and Yang, Dong and Myronenko, Andriy and Landman, Bennett and Roth, Holger R. and Xu, Daguang},
  title={{UNETR}: Transformers for 3D medical image segmentation},
  booktitle={IEEE Winter Conference on Applications of Computer Vision},
  pages={574--584},
  year={2022}
}

@article{oquab2024dinov2,
  author={Oquab, Maxime and others},
  title={{DINOv2}: Learning robust visual features without supervision},
  journal={Transactions on Machine Learning Research},
  year={2024}
}

@article{muller2025mst,
  author={M{\"u}ller-Franzes, Gustav and others},
  title={Medical Slice Transformer for improved diagnosis and explainability on 3D medical images with {DINOv2}},
  journal={Scientific Reports},
  volume={15},
  number={1},
  pages={23979},
  year={2025},
  doi={10.1038/s41598-025-09041-8}
}

@inproceedings{tu2022maxvit,
  author={Tu, Zhengzhong and others},
  title={{MaxViT}: Multi-axis vision transformer},
  booktitle={European Conference on Computer Vision},
  pages={459--479},
  year={2022},
  doi={10.1007/978-3-031-20053-3_27}
}

@inproceedings{chen2021simsiam,
  author={Chen, Xinlei and He, Kaiming},
  title={Exploring simple Siamese representation learning},
  booktitle={Proceedings of the IEEE/CVF Conference on Computer Vision and Pattern Recognition},
  pages={15750--15758},
  year={2021}
}

@article{chen2020moco,
  author={Chen, Xinlei and Fan, Haoqi and Girshick, Ross and He, Kaiming},
  title={Improved baselines with momentum contrastive learning},
  journal={arXiv preprint arXiv:2003.04297},
  year={2020}
}

@article{tejani2024claim,
  author={Tejani, Ali S. and others},
  title={Checklist for Artificial Intelligence in Medical Imaging ({CLAIM}): 2024 update},
  journal={Radiology: Artificial Intelligence},
  volume={6},
  number={4},
  pages={e240300},
  year={2024},
  doi={10.1148/ryai.240300}
}

@inproceedings{chiu2026volumes,
  author={Chiu, Po-Kai and Chen, Hung-Hsuan},
  title={From Volumes to Slices: Computationally Efficient Contrastive Learning for Sequential Abdominal {CT} Analysis},
  booktitle={IEEE International Symposium on Biomedical Imaging},
  pages={1--4},
  year={2026}
}

@inproceedings{wu2022associated,
  author={Wu, Dennis Y. and Lin, Di-Nan and Chen, Vincent F. and Chen, Hung-Hsuan},
  title={Associated Learning: A Methodology to Decompose End-to-End Backpropagation on {CNN}, {RNN}, and Transformer},
  booktitle={International Conference on Learning Representations},
  year={2022},
  url={https://openreview.net/forum?id=4N-17dske79}
}

@article{huang2025deinforeg,
  author={Huang, Zih-Hao and Lin, You-Teng and Chen, Hung-Hsuan},
  title={{DeInfoReg}: A Decoupled Learning Framework with Information Regularization for Better Training Throughput},
  journal={Neurocomputing},
  volume={651},
  pages={130813},
  year={2025},
  doi={10.1016/j.neucom.2025.130813}
}

@article{ho26scpl,
    author = {Ho, Ming-Yao and Wang, Cheng-Kai and Lin, You-Teng and Chen, Hung-Hsuan},
    title = {{SCPL}: Enhancing Neural Network Training Throughput with Decoupled Local Losses and Model Parallelism},
    year = {2026},
    issue_date = {June 2026},
    publisher = {Association for Computing Machinery},
    address = {New York, NY, USA},
    volume = {17},
    number = {2},
    issn = {2158-656X},
    doi = {10.1145/3793534},
    journal = {ACM Trans. Manage. Inf. Syst.},
    month = jun,
    articleno = {6},
    numpages = {25},
}
